\documentclass[10pt,twocolumn,letterpaper]{article}

\usepackage[pagenumbers]{cvpr} 

\usepackage[dvipsnames]{xcolor}

\usepackage{graphicx}
\usepackage{amsmath}
\usepackage{amssymb}
\usepackage{booktabs}
\usepackage{tabularx}

\graphicspath{{images}}

\definecolor{cvprblue}{rgb}{0.21,0.49,0.74}
\usepackage[pagebackref,breaklinks,colorlinks,citecolor=cvprblue]{hyperref}

\def\paperID{*****} 
\def\confName{CVPR}
\def\confYear{2024}

\title{Early Fusion of Features for Semantic Segmentation}

\author{Anupam Gupta \qquad Ashok Krishnamurthy \qquad Lisa Singh \\
Department of Computer Science, University of Virginia\\
{\tt\small angupta@virginia.edu}
}

\begin{document}

\maketitle

\begin{abstract}
This paper introduces a novel segmentation framework that integrates a classifier network with a reverse HRNet architecture for efficient image segmentation. Our approach utilizes a ResNet-50 backbone, pretrained in a semi-supervised manner, to generate feature maps at various scales. These maps are then processed by a reverse HRNet, which is adapted to handle varying channel dimensions through $1 \times 1$ convolutions, to produce the final segmentation output. We strategically avoid fine-tuning the backbone network to minimize memory consumption during training. Our methodology is rigorously tested across several benchmark datasets including Mapillary Vistas, Cityscapes, CamVid, COCO, and PASCAL-VOC2012, employing metrics such as pixel accuracy and mean Intersection over Union (mIoU) to evaluate segmentation performance. The results demonstrate the effectiveness of our proposed model in achieving high segmentation accuracy, indicating its potential for various applications in image analysis. By leveraging the strengths of both the ResNet-50 and reverse HRNet within a unified framework, we present a robust solution to the challenges of image segmentation.
\end{abstract}

\begin{figure*}
\begin{center}
\includegraphics[width=1.0\linewidth]{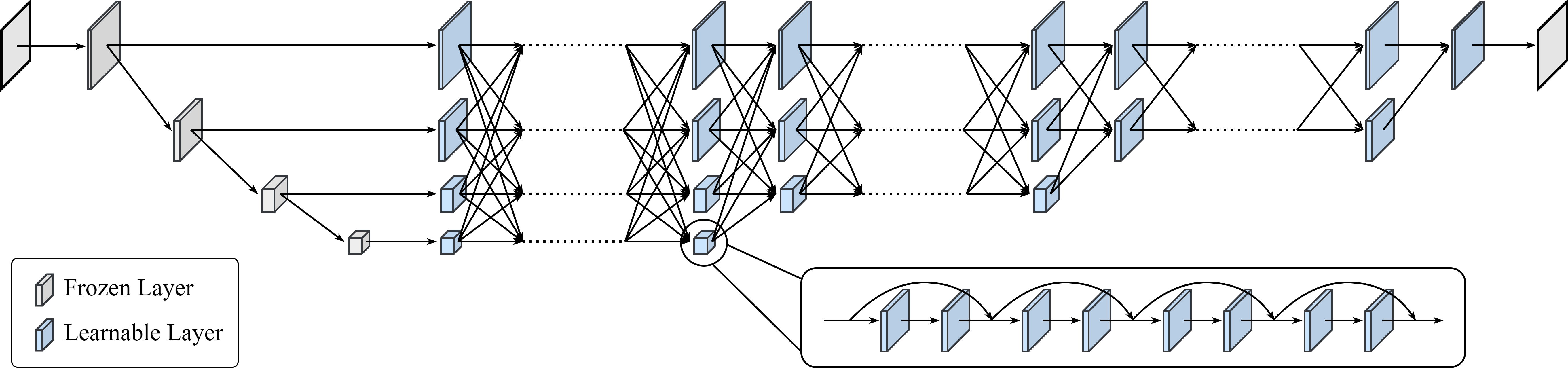}
\end{center}
\caption{The proposed network. The gray blocks are the pretrained backbone and are kept frozen during training.}
\label{fig:network}
\end{figure*}

\section{Introduction}
Deep convolutional neural networks (DCNNs) have emerged as a cornerstone in the advancement of computer vision, excelling in a wide array of tasks including image classification, object detection, semantic segmentation, and human pose estimation, among others. Semantic segmentation, in particular, involves the intricate process of classifying each pixel within an image into distinct class labels, thereby providing a granular and comprehensive scene understanding by identifying the label, location, and shape of every element in the image. This technology holds immense potential in various applications such as autonomous driving and robotic sensing.
\par
Among the plethora of deep learning architectures that have been developed, High-Resolution Network (HRNet) stands out as a notable contribution for its exceptional performance in maintaining high-resolution representations throughout the model \cite{wang2019deep}. Unlike conventional approaches that primarily focus on recovering high-resolution features from a predominantly low-resolution output of a classification network, HRNet innovatively maintains high-resolution streams throughout the entire network. This is particularly advantageous for tasks that are sensitive to spatial details, such as semantic segmentation, human pose estimation, and object detection.
\par
HRNet achieves this by executing convolutions across multiple scales in parallel streams, thereby enriching the model with high-level semantic features without sacrificing low-level details. It incorporates a unique fusion mechanism that allows each stream to integrate features from all other streams, ensuring a comprehensive feature representation at every scale. The architecture is designed to start processing with a high-resolution image, progressively introducing lower resolution streams at each subsequent stage. This progression from high to lower resolutions across stages, from a singular stream in the initial stage to four streams in the fourth and final stage, enables a balanced capture of features across different resolutions.
\par
This approach not only capitalizes on the strengths of maintaining high-resolution representations for accuracy in spatially sensitive tasks but also sets a new benchmark in the efficiency and effectiveness of semantic segmentation and related computer vision applications.
\par
In this study, we introduce an innovative adaptation of the High-Resolution Network (HRNet) by developing an encoder-decoder structure through the concatenation of a reverse HRNet to its standard form. This modification enables us to leverage the HRNet architecture as the encoder component within an encoder-decoder framework, which is traditionally renowned for its efficacy in semantic segmentation tasks. However, implementing a full HRNet-based encoder-decoder model poses significant memory demands during the training phase. To mitigate this challenge without compromising the integrity of low-resolution stream accuracy, we propose the integration of a pre-trained deep convolutional classifier as the encoder and a reverse HRNet as the decoder.
\par
Our model is rigorously evaluated across several benchmark datasets for semantic segmentation, including Mapillary Vistas \cite{neuhold2017mapillary}, Cityscapes \cite{cordts2016cityscapes}, CamVid \cite{BrostowFC:PRL2008}, COCO \cite{lin2014microsoft}, and PASCAL-VOC2012 \cite{pascal-voc-2012}. A key aspect of our investigation focuses on the impact of introducing an additional stream into the HRNet architecture. In a novel approach, we modify the standard HRNet input stem to reduce the resolution to $1/2$ instead of $1/4$, incorporating an extra stream that operates at a resolution higher than the existing streams. This adjustment aims to enhance the network's capability to process high-resolution information without increasing the overall memory footprint. To achieve this balance, the redesigned network features a reduced number of modules at each stage, ensuring that memory consumption remains comparable to the original model.
\par
This strategic modification not only addresses the memory constraints associated with high-resolution semantic segmentation tasks but also explores the potential benefits of higher-resolution processing streams in improving the accuracy and detail of segmentation outcomes.

\section{Related Work}
Semantic segmentation, a critical task in computer vision, has seen significant advancements with the advent of convolutional neural networks (CNNs) embodying various architectural paradigms. Among these, encoder-decoder and hourglass structures have been widely adopted for applications including object detection \cite{lin2017feature,khoshsirat2023improving,khoshsirat2023transformer}, human pose estimation \cite{newell2016stacked,khoshsirat2023sentence,khoshsirat2022semantic}, image-based localization \cite{melekhov2017image,khoshsirat2023empowering,hosseini2022application}, and semantic segmentation itself \cite{long2015fully, badrinarayanan2017segnet, noh2015learning}. These architectures are characterized by an encoder that diminishes the spatial dimensions of feature maps while encapsulating high-level semantic information, and a decoder that progressively restores spatial resolution and detail.
\par
A common challenge faced by these networks is the loss of fine image details during the encoding phase, which hampers achieving optimal results without interventions such as skip connections. The U-net architecture \cite{ronneberger2015u} effectively addresses this by reintroducing encoder features to the decoder through skip connections, enhancing the recovery of low-level details. However, simply deepening the network with more convolutional layers does not necessarily lead to significant accuracy gains.
\par
Spatial pyramid pooling architectures, such as PSPNet \cite{zhao2017pyramid} and DeepLab \cite{chen2017rethinking}, introduce another strategy by implementing pyramid pooling or parallel atrous convolutions at various scales to capture context at multiple resolutions. DeepLabv3+ \cite{chen2018encoder} further refines this approach by incorporating a skip connection to preserve some low-level details from earlier in the network.
\par
High-resolution representation networks represent a different approach, striving to maintain high-resolution feature maps throughout the network to ensure both high-level and low-level details are preserved \cite{wang2019deep,huang2017multi,fourure2017residual,zhou2015interlinked}. These networks perform parallel convolutions at multiple resolutions, effectively balancing the capture of detailed and semantic features. However, they typically require significant memory resources and often necessitate initial downsampling of the input to manage computational demands.
\par
Moreover, some methods enhance segmentation precision, particularly around object edges, by applying post-processing techniques like conditional random fields (CRFs) to the network's outputs \cite{chen2017deeplab,chandra2016fast,khoshsiratembedding,maserat201743}. While effective in refining segmentation details, these techniques introduce additional computational overhead during both training and inference phases.
\par
In summary, the evolution of CNN architectures for semantic segmentation has led to a variety of innovative solutions aimed at overcoming the inherent trade-offs between high-level semantic understanding and low-level detail preservation. Each approach presents its unique advantages and challenges, highlighting the dynamic and multifaceted nature of research in this domain.

\section{Method}
In this work, we introduce a novel two-component architecture designed to enhance the performance of image segmentation tasks. The foundational layer of our model employs the ResNet-50 \cite{he2016deep}, adapted through semi-supervised training techniques as delineated in \cite{yalniz2019billion}, to serve as a classifier network. This initial phase of our architecture is tasked with generating four distinct feature maps, each at varying scales, to capture a comprehensive range of image details and semantic information.
\par
Building upon this, the second segment of our network incorporates a reverse High-Resolution Network (HRNet) \cite{wang2019deep}. This innovative adaptation is specifically engineered to accept the quartet of feature maps produced by the initial classifier network, facilitating a seamless transition from feature extraction to the final stage of image segmentation. The reverse HRNet's role is critical, as it synthesizes the input feature maps to produce the final segmented image, a process elegantly illustrated in Figure \ref{fig:network}.
\par
Given the potential discrepancy in the number of channels across corresponding streams between the two network components, we employ $1 \times 1$ convolutions to standardize the channel count, ensuring a harmonious integration of the classifier and reverse HRNet outputs. Furthermore, to optimize memory efficiency during the training phase, we refrain from fine-tuning the backbone network. This strategic decision aids in managing computational resources without compromising the integrity of the segmentation outcomes.
\par
To facilitate a direct and fair comparison with the original HRNet model, our experimental setup adheres to the identical hyper-parameters and configuration settings as established in \cite{wang2019deep}. This approach ensures that any observed performance enhancements can be attributed to the structural innovations of our proposed network, thereby providing a clear benchmark for evaluating its effectiveness in the realm of image segmentation.

\newcommand{\addimg}[1]{\includegraphics[width=0.15\textwidth]{#1.jpg}}
\begin{figure*}[t]
\centering
\begin{tabular}{cccccc}
\hline
Method & Prediction & Stride-4 & Stride-8 & Stride-16 & Stride-32 \\
\hline
HRNet & \addimg{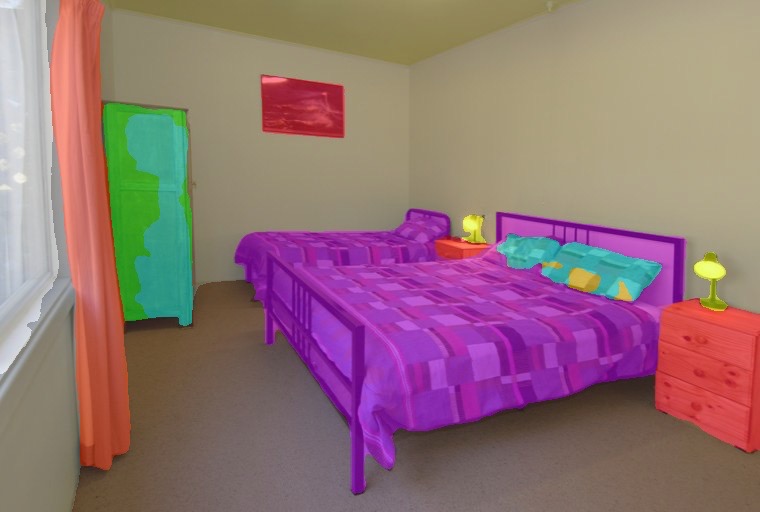} & \addimg{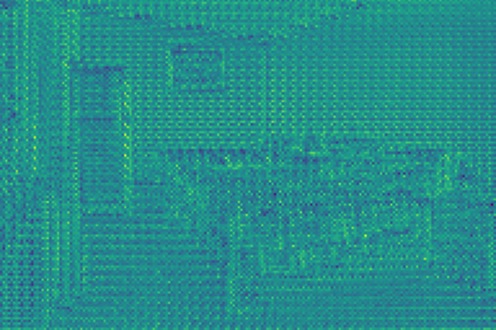} & \addimg{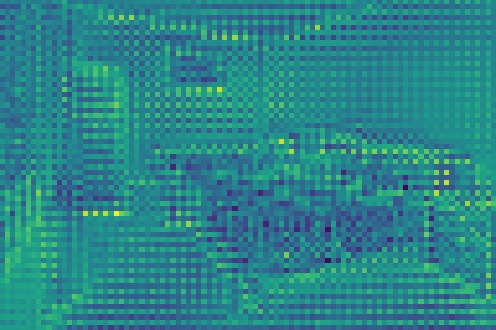} & \addimg{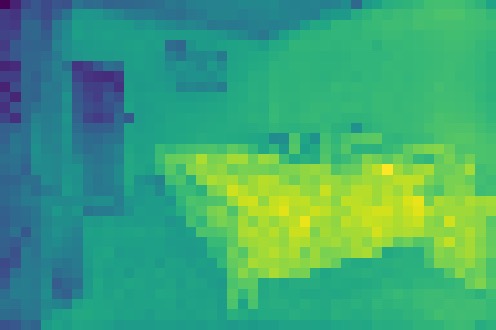} & \addimg{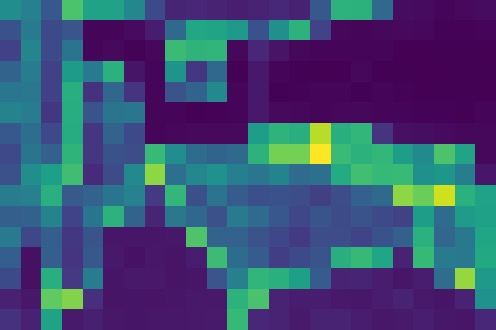} \\
\textbf{Ours} & \addimg{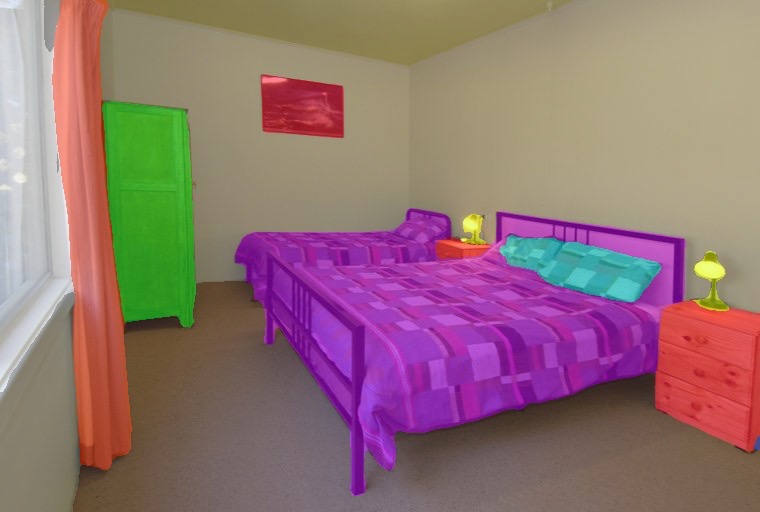} & \addimg{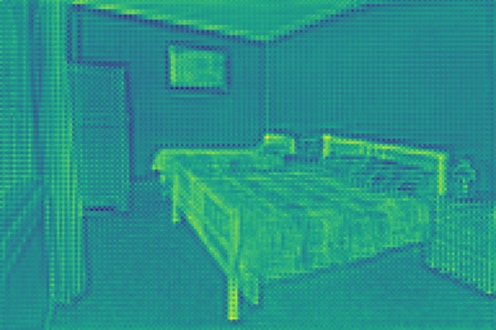} & \addimg{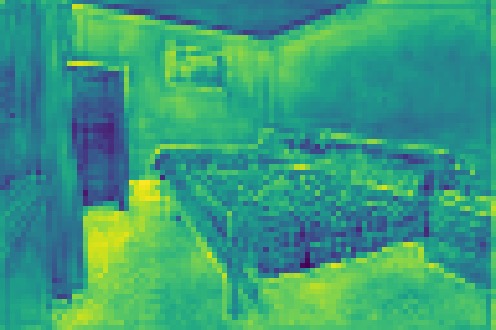} & \addimg{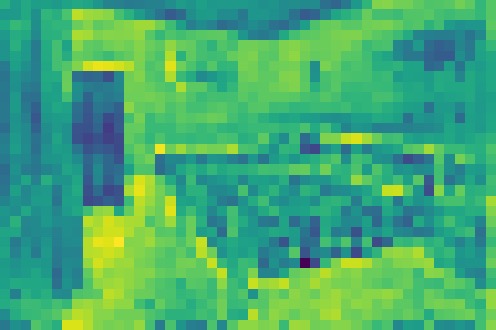} & \addimg{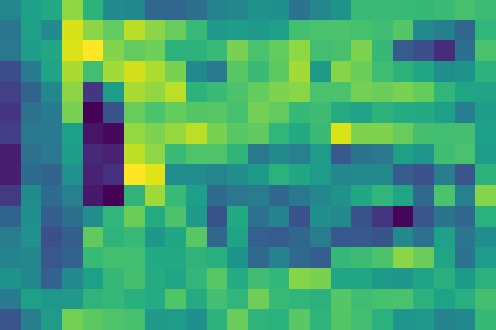} \\
\hline
HRNet & \addimg{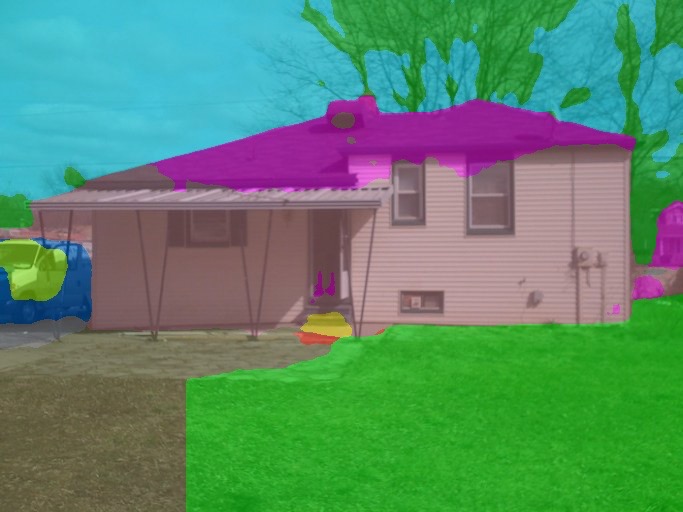} & \addimg{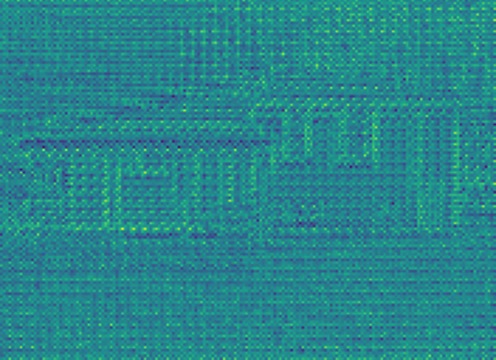} & \addimg{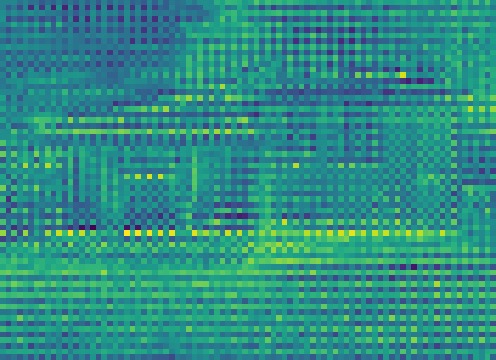} & \addimg{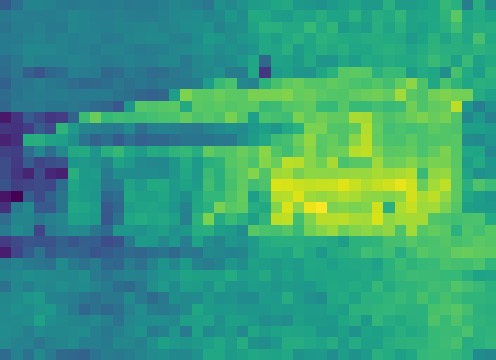} & \addimg{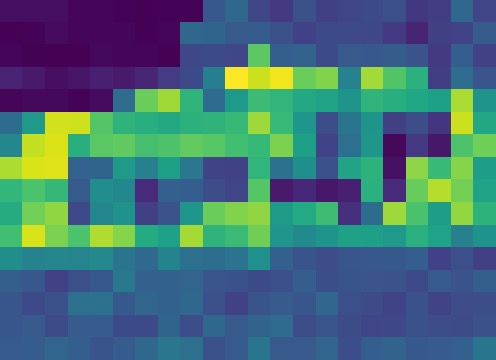} \\
\textbf{Ours} & \addimg{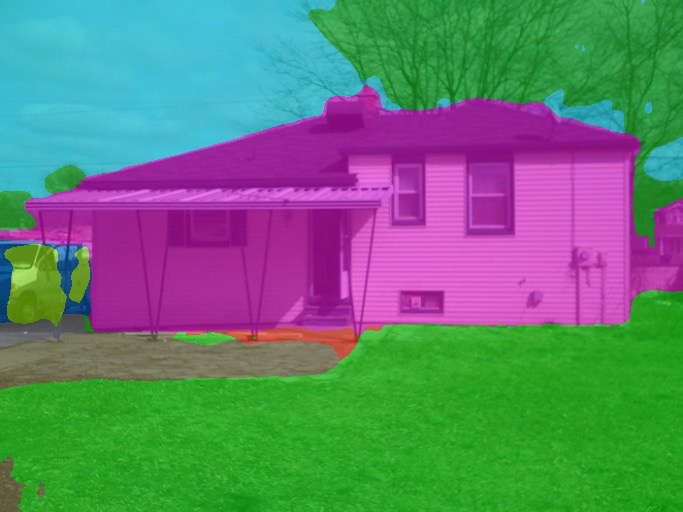} & \addimg{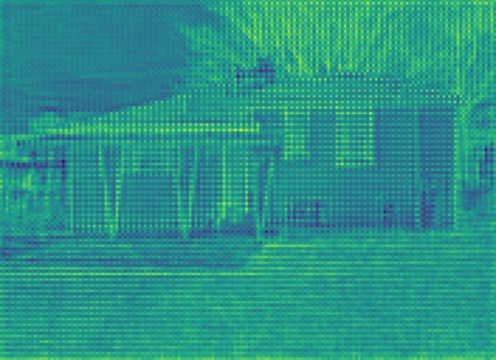} & \addimg{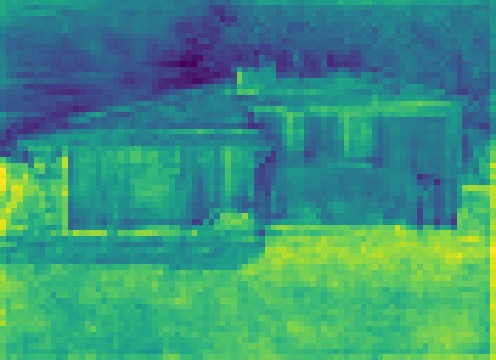} & \addimg{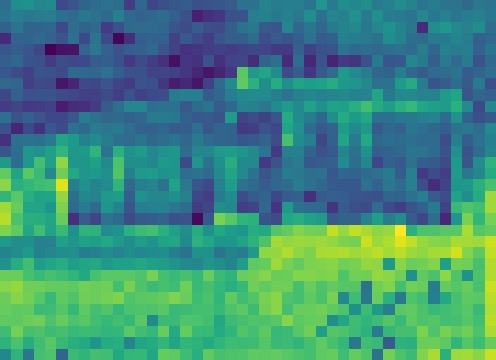} & \addimg{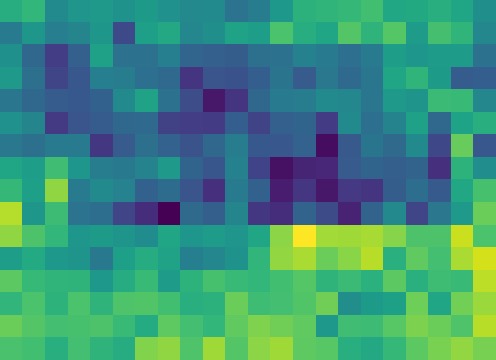} \\
\hline
HRNet & \addimg{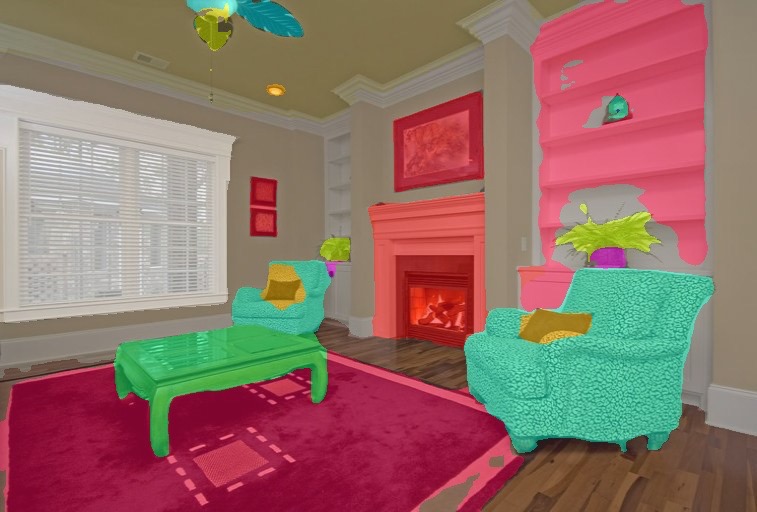} & \addimg{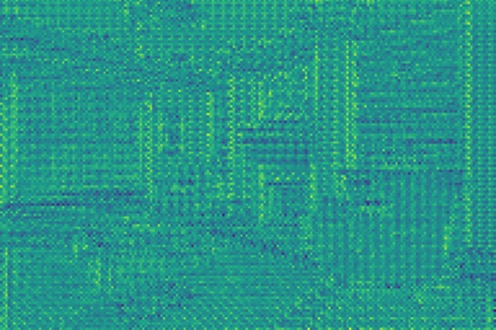} & \addimg{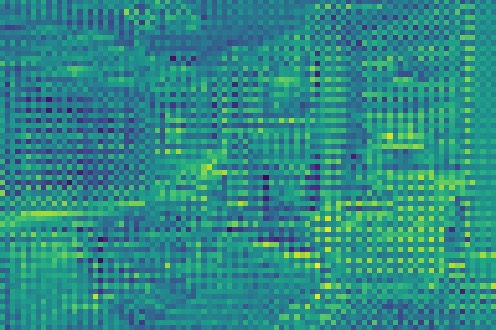} & \addimg{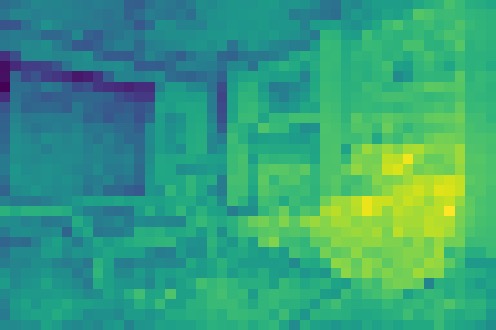} & \addimg{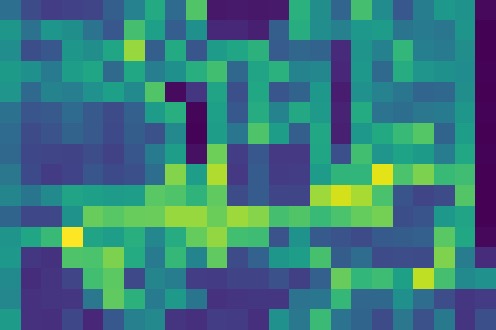} \\
\textbf{Ours} & \addimg{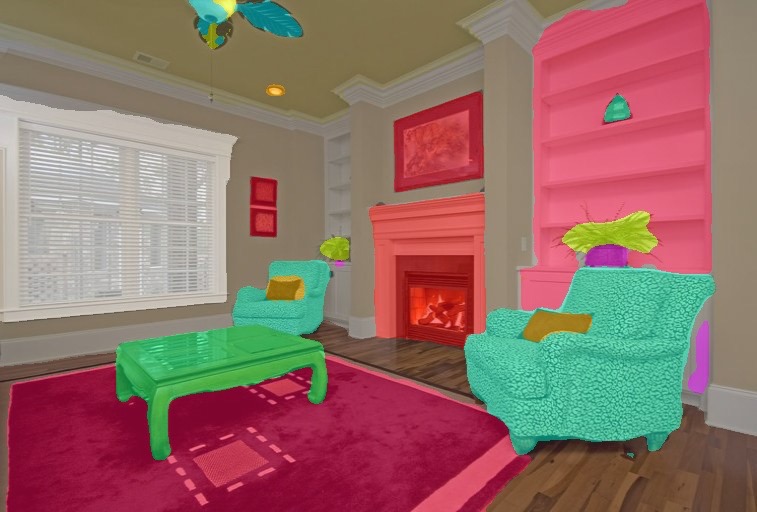} & \addimg{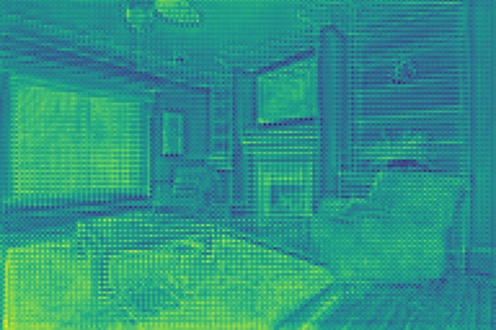} & \addimg{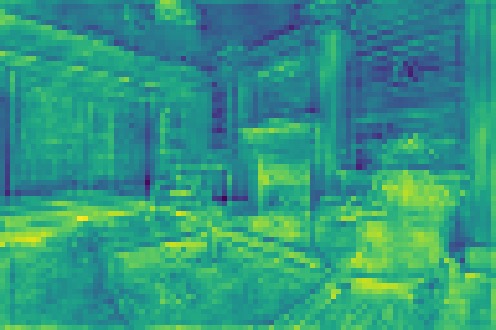} & \addimg{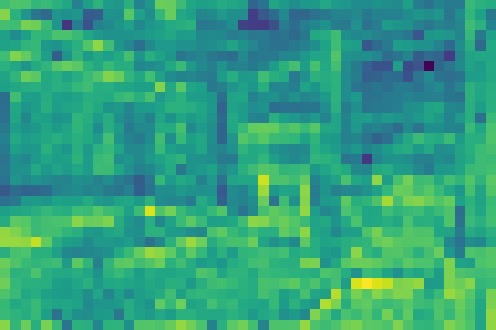} & \addimg{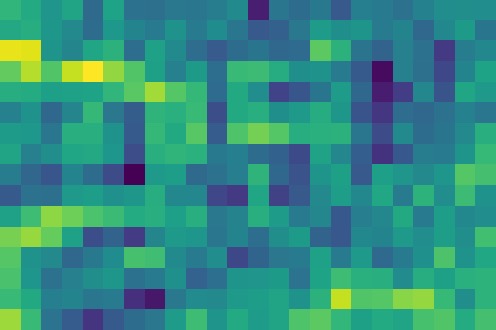} \\
\hline
HRNet & \addimg{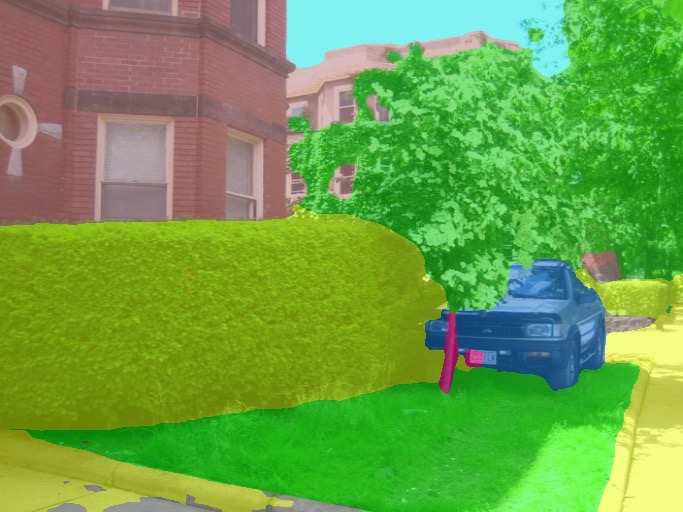} & \addimg{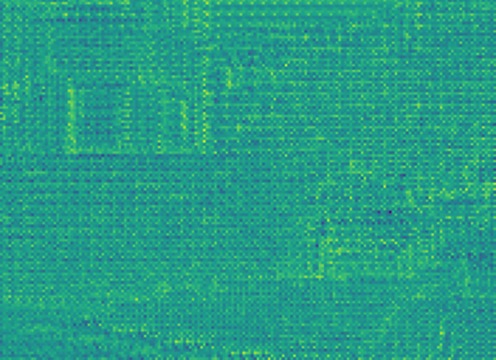} & \addimg{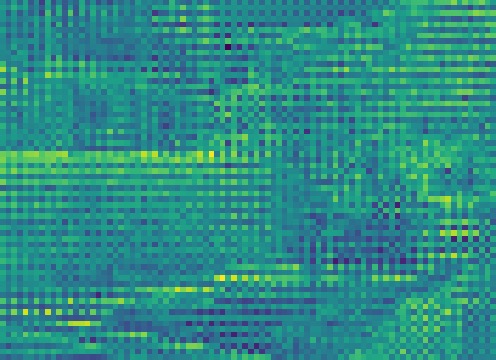} & \addimg{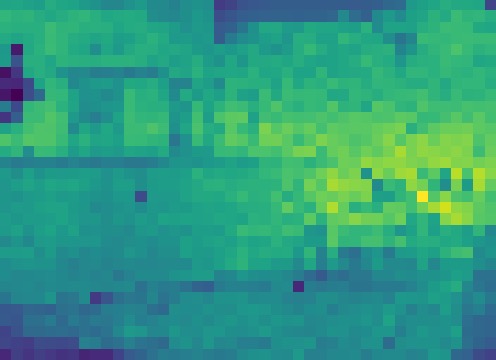} & \addimg{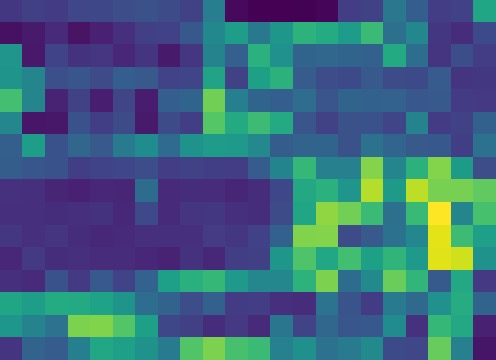} \\
\textbf{Ours} & \addimg{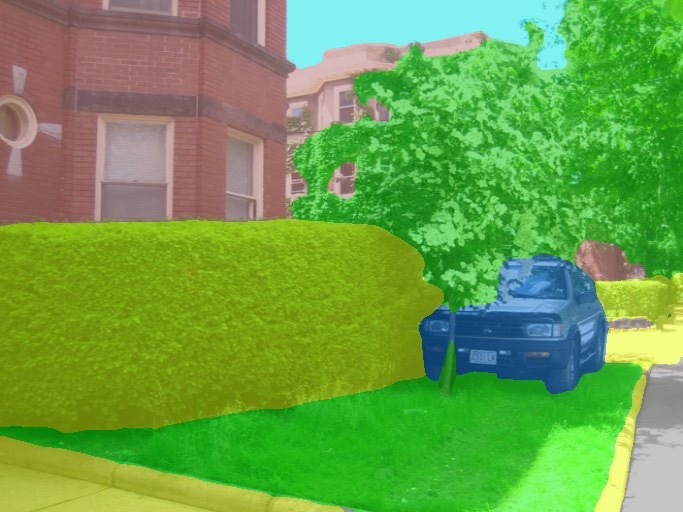} & \addimg{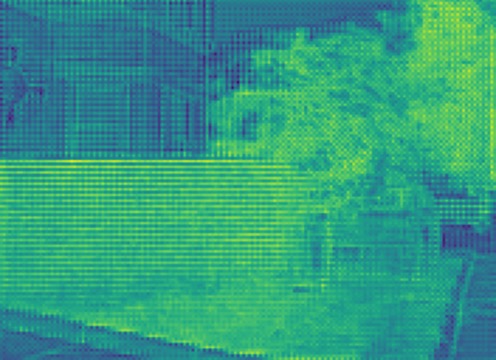} & \addimg{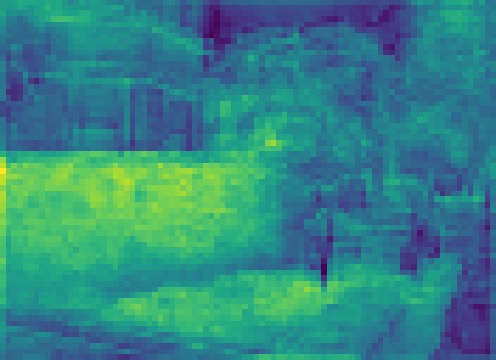} & \addimg{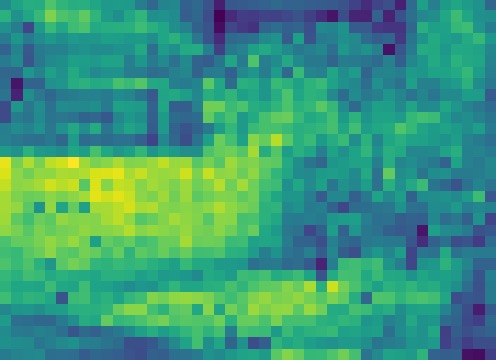} & \addimg{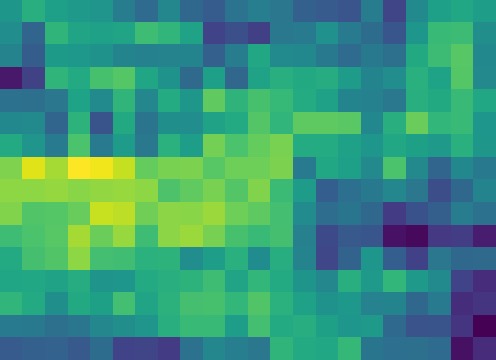} \\
\hline
HRNet & \addimg{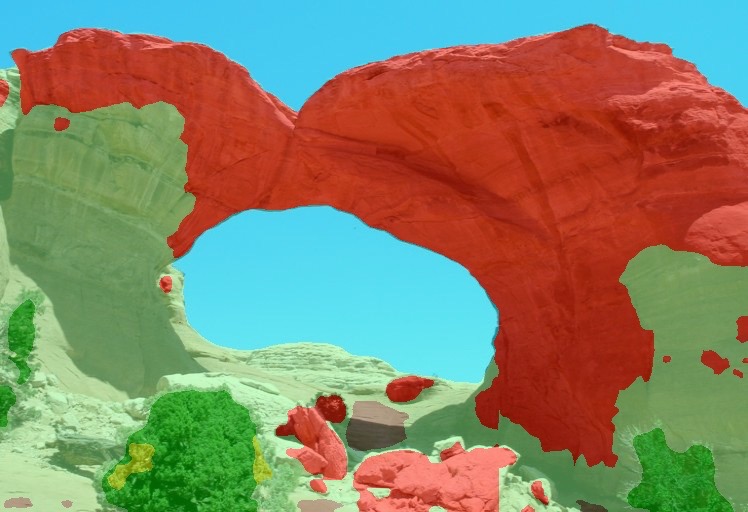} & \addimg{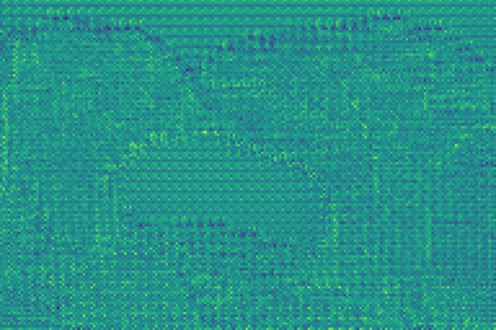} & \addimg{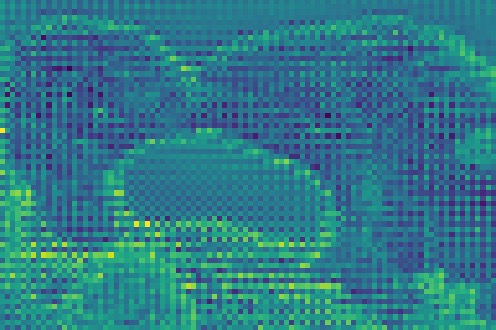} & \addimg{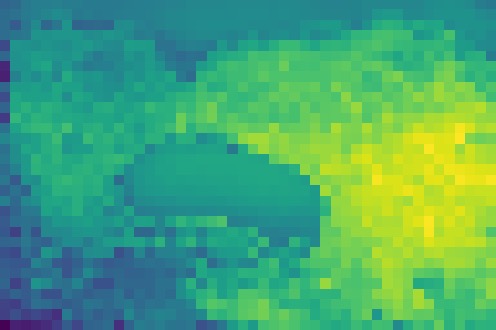} & \addimg{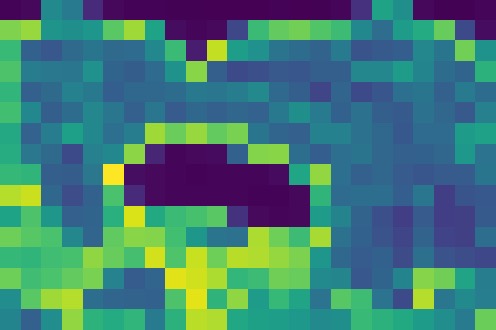} \\
\textbf{Ours} & \addimg{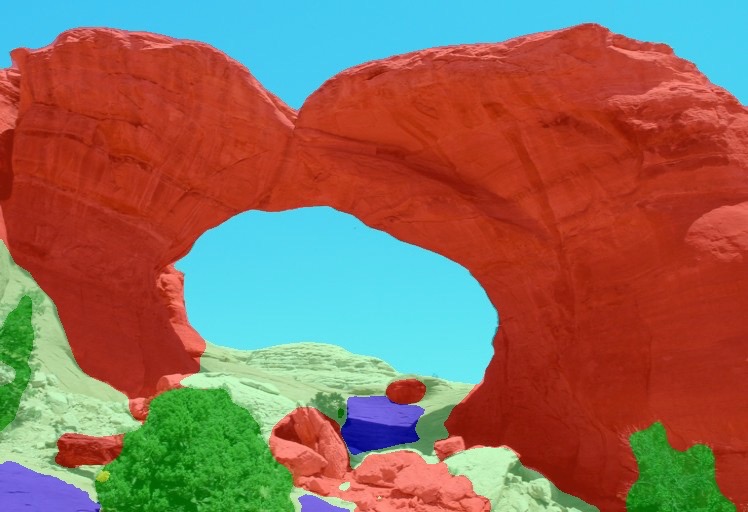} & \addimg{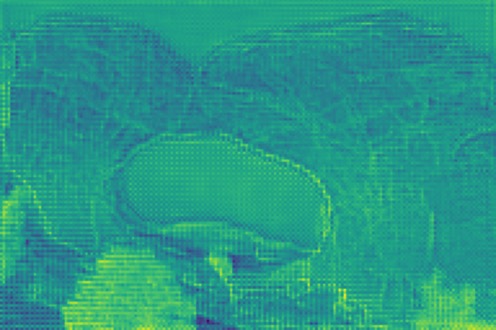} & \addimg{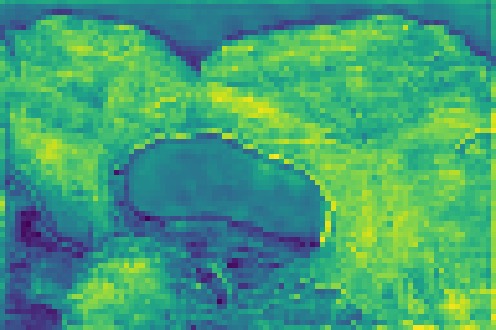} & \addimg{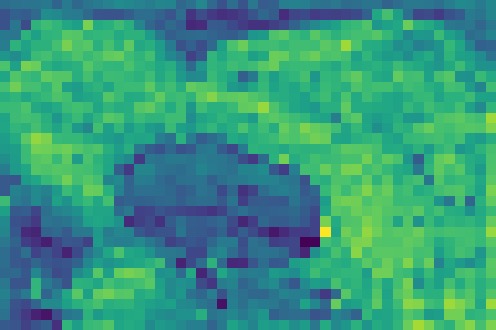} & \addimg{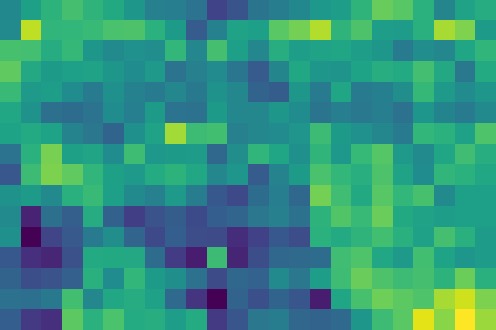} \\
\hline
\end{tabular}
\caption{This figure presents a detailed visual comparison of semantic segmentation results between our proposed method and HRNet. Demonstrating the advantages of our approach, it is evident that our method produces more refined multi-scale feature maps, capturing intricate local edges and textures with superior precision. This enhancement in feature map quality significantly contributes to the improved performance of semantic segmentation, showcasing our method's ability to handle complex scenes with enhanced detail and accuracy.}
\label{fig:visual-results}
\end{figure*}

\section{Experiments}
To validate the efficacy of our proposed network architecture, we embarked on a comprehensive training and testing regimen leveraging multiple benchmark datasets renowned within the domain of semantic segmentation. Initially, we pretrain our model using the extensive Mapillary Vistas dataset \cite{neuhold2017mapillary}, subsequently applying this pretrained model to both the Cityscapes \cite{cordts2016cityscapes} and CamVid \cite{BrostowFC:PRL2008} datasets through further training and evaluation phases. This sequential approach allows us to refine the model's capabilities in diverse urban environments.
\par
In parallel, we employ a similar pretraining strategy with the COCO dataset \cite{lin2014microsoft}, renowned for its broad array of object instances, followed by dedicated training and testing on the PASCAL-VOC2012 dataset \cite{pascal-voc-2012}. This dual-path pretraining strategy is designed to imbue our model with a robust understanding of both large-scale environmental contexts and detailed object classifications.
\par
Given the existing pretraining of the backbone network, we consciously decide against additional pretraining on the ImageNet dataset \cite{russakovsky2015imagenet}. This decision is motivated by our intention to streamline the training process, focusing on datasets that offer the most direct relevance to the task of semantic segmentation.
\par
Throughout all phases of experimentation, we employ two key metrics to assess the segmentation performance of our model: pixel accuracy and the mean Intersection over Union (mIoU). These metrics are chosen for their widespread acceptance and ability to provide a nuanced understanding of model performance, encompassing both the overall accuracy and the precision of the segmentation boundaries. Through this methodical approach, we aim to thoroughly evaluate the proposed model's capabilities across a variety of challenging visual contexts.

\section{Datasets}
This section provides an overview of the datasets employed in our study to evaluate the performance of the proposed network architecture for semantic segmentation tasks. Each dataset is selected for its unique characteristics and relevance to the field, offering a comprehensive assessment across various scenarios.

\subsection{Mapillary Vistas}
The Mapillary Vistas research edition \cite{neuhold2017mapillary} is a vast collection of street-level images, encompassing 25,000 densely annotated images. These images are distributed across training, validation, and test sets with counts of 18,000, 2,000, and 5,000, respectively. The dataset features 65 object categories alongside a void class, with image resolutions reaching up to 22 Megapixels and varying aspect ratios, providing a challenging and diverse set for semantic segmentation.

\subsection{Cityscapes}
Cityscapes \cite{cordts2016cityscapes} offers 5,000 high-quality, pixel-level annotated images specifically tailored for urban street scenes. The dataset is organized into sets of 2,975 training, 500 validation, and 1,525 test images, complemented by an additional 20,000 coarsely annotated images. Evaluation is conducted on 19 of the 30 available classes, using metrics including mean Intersection over Union (mIoU), IoU category, iIoU class, and iIoU category, to provide a thorough analysis of model performance.

\subsection{CamVid}
The CamVid dataset \cite{BrostowFC:PRL2008} offers a focused examination of driving scenarios through a smaller set of 701 annotated images derived from five video sequences, categorized into 32 classes and resized to 960$\times$720. Following the protocol introduced by \cite{badrinarayanan2017segnet}, we segment this dataset into 367 training, 101 validation, and 233 test images across 11 classes, allowing for a focused study on semantic segmentation in automotive contexts.

\subsection{COCO}
The comprehensive COCO dataset \cite{lin2014microsoft} encompasses over 123,000 images for training and validation, spread across 80 object categories. This dataset is known for its variety in image sizes and aspect ratios, offering a broad spectrum of object instances and scenes for segmentation tasks.

\subsection{PASCAL-VOC2012}
The PASCAL-VOC2012 \cite{pascal-voc-2012} segmentation dataset includes images across 20 object categories plus one background class. With 1,465 training, 1,450 validation, and 1,456 test images, augmented by an additional 10,582 training images from \cite{hariharan2011semantic}, this dataset provides a rigorous benchmark for evaluating segmentation accuracy.

\section{Conclusion}
The development and evaluation of our novel segmentation framework, integrating a classifier network with a reverse HRNet architecture, have yielded promising results across multiple benchmark datasets. This approach leverages the strength of ResNet-50 as a feature extractor, combined with the sophisticated segmentation capabilities of the reverse HRNet, to produce high-quality image segmentation. Our methodology's effectiveness is underscored by rigorous testing on diverse datasets such as Mapillary Vistas, Cityscapes, CamVid, COCO, and PASCAL-VOC2012, employing metrics like pixel accuracy and mean Intersection over Union (mIoU) to assess performance.
\par
Our research confirms the critical role of high-resolution feature maps in enhancing segmentation accuracy, particularly in tasks requiring precise spatial detail. By maintaining high-resolution streams throughout the network, our model effectively captures both high-level semantic and low-level detail information, a balance that is often challenging to achieve. This balance is further optimized by our strategic decision not to fine-tune the backbone network, thereby reducing memory consumption without compromising the quality of segmentation results.
\par
Moreover, the introduction of an additional stream, specifically designed to process higher-resolution information, has demonstrated a significant impact on the model's performance. This modification, coupled with the use of a pre-trained deep convolutional classifier as the encoder and a tailored reverse HRNet as the decoder, presents a novel approach to handling the computational and memory demands typically associated with high-resolution image segmentation.
\par
The extensive evaluation conducted as part of this study not only validates the effectiveness of our proposed model but also highlights its potential applicability in real-world scenarios where high-quality segmentation is critical. Despite the limitations related to memory consumption and computational efficiency, our findings suggest that the strategic modifications introduced in our network architecture significantly contribute to improving segmentation outcomes.
\par
In conclusion, our work presents a compelling case for the integration of high-resolution feature maps and advanced network structures in addressing the challenges of semantic segmentation. Future work will aim to further refine this approach, exploring additional strategies for reducing computational demands while enhancing model accuracy and efficiency.

{
    \small
    \bibliographystyle{ieeenat_fullname}
    \bibliography{main}
}

\end{document}